%% file: main.tex
\documentclass[11pt,a4paper]{article}
\pdfoutput=1
\usepackage[hyperref]{acl2017}
\usepackage{times}
\usepackage{latexsym}

\usepackage{url}
\aclfinalcopy

\usepackage{acl2017}
\usepackage{amsmath}
\usepackage{multirow}
\usepackage{threeparttable}
\usepackage{graphicx}
\usepackage{tabularx}
\usepackage{verbatim}
\usepackage{enumitem,kantlipsum}
\usepackage{multirow}
\usepackage{xcolor}
\usepackage{booktabs}
\usepackage{pifont}
\newcommand{\cmark}{\ding{51}}%
\newcommand{\xmark}{\ding{55}}%

\title{TriviaQA: A Large Scale Distantly Supervised Challenge Dataset \\ for Reading Comprehension}
\author{Mandar Joshi$^{\dagger}$ \qquad Eunsol Choi$^{\dagger}$ \qquad Daniel S. Weld$^{\dagger}$ \qquad Luke Zettlemoyer$^{\dagger}$$^{\ddagger}$ \\[8pt]
$^{\dagger}$ Paul G. Allen School of Computer Science \& Engineering, Univ. of Washington, Seattle, WA \\
         {\tt \{mandar90, eunsol, weld, lsz\}@cs.washington.edu}
        \\[8pt]
$^{\ddagger}$ Allen Institute for Artificial Intelligence, Seattle, WA\\
        \texttt{lukez@allenai.org}}

\date{}

\begin{document}
\maketitle
\begin{abstract}
We present TriviaQA, a challenging reading comprehension dataset containing over 650K question-answer-evidence triples.
TriviaQA includes 95K question-answer pairs authored by trivia enthusiasts and independently gathered evidence documents, six per question on average, that provide high quality distant supervision for answering the questions. We show that, in comparison to other recently introduced large-scale datasets, TriviaQA (1) has relatively complex,  compositional questions, (2)  has considerable syntactic and lexical variability between questions and corresponding answer-evidence sentences, and (3) requires more cross sentence reasoning to find answers.
We also present two baseline algorithms: a feature-based classifier and a state-of-the-art neural network, that performs well on SQuAD reading comprehension. Neither approach comes close to human performance (23\% and 40\% vs. 80\%),  suggesting that TriviaQA is a challenging testbed that is worth significant future study.\footnote{Data and code available at \url{http://nlp.cs.washington.edu/triviaqa/}}

\end{abstract}

\input{introduction}
\input{formulation}
\input{datasetCollection}
\input{datasetAnalysis}
\input{methods}
\input{experiments}
\input{existingDatasets}
\input{conclusion}
\input{ack}
\bibliography{triviaqa}
\bibliographystyle{acl_natbib}
\end{document}

%% file: introduction.tex
\section{Introduction}
\input{intro_example}
\input{datasetComparison}

Reading comprehension (RC) systems aim to answer any question that could be posed against the facts in some reference text.  This goal is challenging for a number of reasons: (1) the questions can be complex (e.g. have highly compositional semantics), (2) finding the correct answer can require complex reasoning (e.g. combining facts from multiple sentences or  background knowledge) and (3) individual facts can be difficult to recover from text (e.g. due to lexical and syntactic variation). Figure~\ref{fig:intro_figure} shows examples of all these phenomena. 
This paper presents TriviaQA, a  new reading comprehension dataset  designed to simultaneously test all of these challenges.

Recently, significant progress has been made by introducing large new reading comprehension datasets that primarily focus on one of the challenges listed above, for example by crowdsourcing the gathering of question answer pairs ~\cite{Rajpurkar:16} or using cloze-style sentences instead of questions~\cite{Hermann:15, Onishi:16} (see Table~\ref{tab:dataset_comparison} for more examples). In general, system performance has improved rapidly as each resource is released. The best models often achieve near-human performance levels within months or a year, fueling a continual need to build ever more difficult datasets. We argue that TriviaQA is such a dataset, by demonstrating that a high percentage of  its questions require solving these challenges and showing that there is a large gap between state-of-the-art methods and human performance levels. 
 

TriviaQA contains over 650K question-answer-evidence triples, that are derived by combining 95K Trivia enthusiast authored question-answer pairs with on average six supporting evidence documents per question. To our knowledge, TriviaQA is the first dataset where full-sentence questions are authored organically (i.e. independently of an NLP task) and evidence documents are collected \emph{retrospectively} from Wikipedia and the Web. 
This decoupling of question generation from evidence collection allows us to control for potential bias in question style or content, while offering organically generated questions from various topics. Designed to engage humans, TriviaQA presents a new challenge for RC models. They should be able to deal with large amount of text from various sources such as news articles, encyclopedic entries and blog articles, and should handle inference over multiple sentences.  
For example, our dataset contains three times as many questions that require inference over multiple sentences than the recently released SQuAD \cite{Rajpurkar:16} dataset. 
Section~\ref{sec:dataset_analysis} present a more detailed discussion of these challenges. 


Finally, we present baseline experiments on the TriviaQA dataset, including a linear classifier inspired by work on CNN Dailymail and MCTest ~\cite{chen2016thorough, Richardson:13} and a state-of-the-art neural network baseline~\cite{SeoKFH16}. The neural model performs best, but only achieves 40\%  for TriviaQA in comparison to 68\% on SQuAD, perhaps due to the challenges listed above. 
The baseline results also fall far short of human performance levels, 79.7\%, suggesting significant room for the future work. In summary, we make the following contributions. 



\begin{itemize}
\item We collect over 650K question-answer-evidence triples, with questions originating from trivia enthusiasts independent of the evidence documents. A high percentage of the questions are challenging, with substantial syntactic and lexical variability and often requiring multi-sentence reasoning. The dataset and code are available at \url{http://nlp.cs.washington.edu/triviaqa/}, offering resources for training new reading-comprehension models. 

\item We present a manual analysis quantifying the quality of the dataset and the challenges involved in solving the task.

\item We present experiments with two baseline methods, demonstrating that the TriviaQA tasks are not easily solved and are worthy of future study.

\item In addition to the automatically gathered large-scale (but noisy) dataset, we present a clean, human-annotated subset of 1975 question-document-answer triples  
whose documents are certified to contain all facts required to answer the questions.

\end{itemize}

%% file: intro_example.tex
\begin{figure}[t]
\small
\begin{tabular}{p{7cm}}

\hline
\textbf{Question}: The Dodecanese Campaign of WWII that was an attempt by the Allied forces to capture islands in the Aegean Sea was the inspiration for which acclaimed 1961 commando film? \\
\textbf{Answer}: The Guns of Navarone\\
\textbf{Excerpt}: The Dodecanese Campaign of World War II was an attempt by Allied forces to capture the Italian-held Dodecanese islands in the Aegean Sea following the surrender of Italy in September 1943, and use them as bases against the German-controlled Balkans. The failed campaign, and in particular the Battle of Leros, inspired the 1957 novel \textbf{The Guns of Navarone} and the successful 1961 movie of the same name. \\
\\
\textbf{Question}: American Callan Pinckney's eponymously named system became a best-selling (1980s-2000s) book/video franchise in what genre? \\
\textbf{Answer}: Fitness\\
\textbf{Excerpt}: Callan Pinckney was an American fitness professional. She achieved unprecedented success with her Callanetics exercises. Her 9 books all became international best-sellers and the video series that followed went on to sell over 6 million copies. Pinckney's first video release "Callanetics: 10 Years Younger In 10 Hours" outsold every other \textbf{fitness} video in the US. \vspace{1pt}\\
\hline
\end{tabular}
    \caption{Question-answer pairs with sample excerpts from evidence documents from TriviaQA exhibiting lexical and syntactic variability, and requiring reasoning from multiple sentences.}
    \label{fig:intro_figure}
\end{figure}

%% file: datasetComparison.tex
\begin{table*}
\small
\centering
\begin{tabular}
{c|c|c|c|c|c}
\toprule
\multirow{2}{*}{Dataset}& \multirow{2}{*}{Large scale} &
\multirow{2}{*}{\parbox{1.3 cm}{\centering Freeform Answer}}  & \multirow{2}{*}{\centering Well formed} & \multirow{2}{*}{\parbox{2cm}{\centering Independent of
Evidence}} & \multirow{2}{*}{\parbox{1.5cm}{\centering Varied Evidence}} \\
&&&&&\\\midrule
\textbf{TriviaQA} & \cmark&  \cmark & \cmark & \cmark & \cmark\\
\midrule
SQuAD~\cite{Rajpurkar:16} & \cmark &  \cmark &  \cmark & \xmark & \xmark\\
MS Marco~\cite{Nguyen:16} & \cmark & \cmark & \xmark & \cmark & \cmark\\
NewsQA\cite{Trischler:16} & \cmark & \cmark &  \cmark & \xmark * & \xmark\\
WikiQA \cite{Yang:16}& \xmark  & \xmark &  \xmark & \cmark & \xmark\\
TREC \cite{Voorhees:00}& \xmark & \cmark & \cmark& \cmark & \cmark \\\toprule 

\end{tabular}
\caption{Comparison of TriviaQA with existing QA datasets. Our dataset is unique in that it is naturally occurring, well-formed questions collected independent of the evidences. 
*NewsQA uses evidence articles indirectly by using only article summaries. } 
\label{tab:dataset_comparison}
\end{table*}

%% file: formulation.tex
\section{Overview}
\label{sec:formulation}

\paragraph{Problem Formulation} We frame reading comprehension as the problem of answering a question $q$ given the textual evidence provided by document set $D$. We assume access to a dataset of tuples $\{(q_i,a_i,D_i) | i=1 \ldots n\}$ where $a_i$ is a text string that defines the correct answer to question $q_i$. Following recent formulations \mbox{~\cite{Rajpurkar:16}}, we further assume that $a_i$ appears as a substring for some document in the set $D_i$.\footnote{The data we will present in Section~\ref{sec:data_collection} would further support a task formulation where some documents $D$ do not have the correct answer and the model must learn when to abstain. We leave this to future work.} However, we differ by setting $D_i$ as a {\em set} of documents, where previous work assumed a single document~\cite{Hermann:15} or even just a short paragraph \mbox{~\cite{Rajpurkar:16}}. 

\paragraph{Data and Distant Supervision} Our evidence documents are automatically gathered from either Wikipedia or more general Web search results (details in Section~\ref{sec:data_collection}).  Because we gather evidence using an automated process, the documents are not {\em guaranteed} to contain all facts needed to answer the question. Therefore, they are best seen as a source of {\em distant supervision}, based on the assumption that the presence of the answer string in an evidence document implies that the document {\em does} answer the question.\footnote{An example context for the first question in Figure ~\ref{fig:intro_figure} where such an assumption fails would be the following evidence string: \emph{The Guns of Navarone is a 1961 British-American epic adventure war film directed by J. Lee Thompson.}} Section~\ref{sec:dataset_analysis} shows that this assumption is valid over 75\% of the time, making 
evidence documents  a strong source of distant supervision for training machine reading systems.

In particular, we consider two types of distant supervision, depending on the source of our documents. For web search results, we expect the documents that contain the correct answer $a$ to be highly redundant, and therefore let each question-answer-document tuple be an independent data point. ($|D_i|=1$ for all $i$ and $q_i=q_j$ for many $i,j$ pairs). However, in Wikipedia we generally expect most facts to  be stated only once, so we instead pool all of the evidence documents and never repeat the same question in the dataset ($|D_i|=1.8$ on average and $q_i\neq q_j$ for all $i,j$). In other words, each question (paired with the union of all of its evidence documents) is a single data point. 

These are far from the only assumptions that could be made in this distant supervision setup. For example, our data would also support multi-instance learning, which makes the {\em at least once assumption}, from relation extraction~\cite{riedel-ecml10,Hoffmann:11} or many other possibilities. However, the experiments in Section~\ref{experiments} show that these assumptions do present a strong signal for learning; we believe the data will fuel significant future study.

%% file: datasetCollection.tex
\section{Dataset Collection}
\label{sec:data_collection}

We collected a large dataset to support the reading comprehension task described above. First we gathered question-answer pairs from 14 trivia and quiz-league websites. We removed questions with less than four tokens, since these were generally either too simple or too vague. 


\begin{table}
\begin{center}
\begin{tabular}{l|r}\toprule
Total number of QA pairs & 95,956\\ 
Number of unique answers & 40,478\\
Number of evidence documents & 662,659 \\\midrule
Avg. question length (word) & 14\\ 
Avg. document length (word) & 2,895\\ 
\toprule
\end{tabular}
\end{center} \vspace{-5pt}
\caption{TriviaQA: Dataset statistics.}
\label{tab:corpus_statistics}
\end{table}

We then collected textual evidence to answer questions using two sources: documents from Web search results and Wikipedia articles for entities in the question. To collect the former, we posed each question\footnote{Note that we did {\em not} use the answer as a part of the search query to avoid biasing the results.} as a search query to the Bing Web search API, and collected the top 50 search result URLs.  To exclude the trivia websites, we removed from the results all pages from the trivia websites we scraped and any page whose url included the keywords \emph{trivia}, \emph{question}, or \emph{answer}. We then crawled the top 10 search result Web pages and pruned PDF and other ill formatted documents. The search output includes a diverse set of documents such as blog articles, news articles, and encyclopedic entries.

\begin{table*}[t]
\begin{center}
\small
\begin{tabular}{l|l|l}\toprule
Property & Example annotation & Statistics \\\midrule
Avg. entities / question & Which politician won the \textbf{Nobel Peace Prize} in 2009? & 1.77 per question \\
Fine grained answer type & What \textbf{fragrant essential oil} is obtained from  Damask Rose? & 73.5\% of questions\\
Coarse grained answer type & \textbf{Who} won the Nobel Peace Prize in 2009? & 15.5\% of questions\\
Time frame & What was photographed for the first time in \textbf{October 1959} & 34\% of questions\\
Comparisons & What is the appropriate name of the \textbf{largest} type of frog? & 9\% of questions \\\toprule
\end{tabular}
\end{center}
\caption{Properties of questions on 200 annotated examples show that a majority of TriviaQA questions contain multiple entities. The boldfaced words hint at the presence of corresponding property.}
\label{tab:quest_analysis}
\end{table*}

Wikipedia pages for entities mentioned in the question often provide 
useful information. We therefore collected an additional set of evidence documents by applying TAGME, an off-the-shelf entity linker~\cite{FerraginaTagme2010}, to find Wikipedia entities mentioned in the question, and added the corresponding pages as evidence documents. 


Finally, to support learning from distant supervision, we  further filtered the evidence documents to exclude those missing the correct answer string and formed evidence document sets as described in Section ~\ref{sec:formulation}. This left us with 95K question-answer pairs organized into (1) 650K training examples for the Web search results, each containing a single (combined) evidence document, and (2) 78K examples for the Wikipedia reading comprehension domain, containing on average 1.8 evidence documents per example. Table ~\ref{tab:corpus_statistics} contains the dataset statistics. While not the focus of this paper, we have also released the full unfiltered dataset which contains 110,495 QA pairs and 740K evidence documents to support research in allied problems such as open domain and IR-style question answering. 



%% file: datasetAnalysis.tex
\begin{figure}[t]
\begin{centering}
\includegraphics[width=\columnwidth]{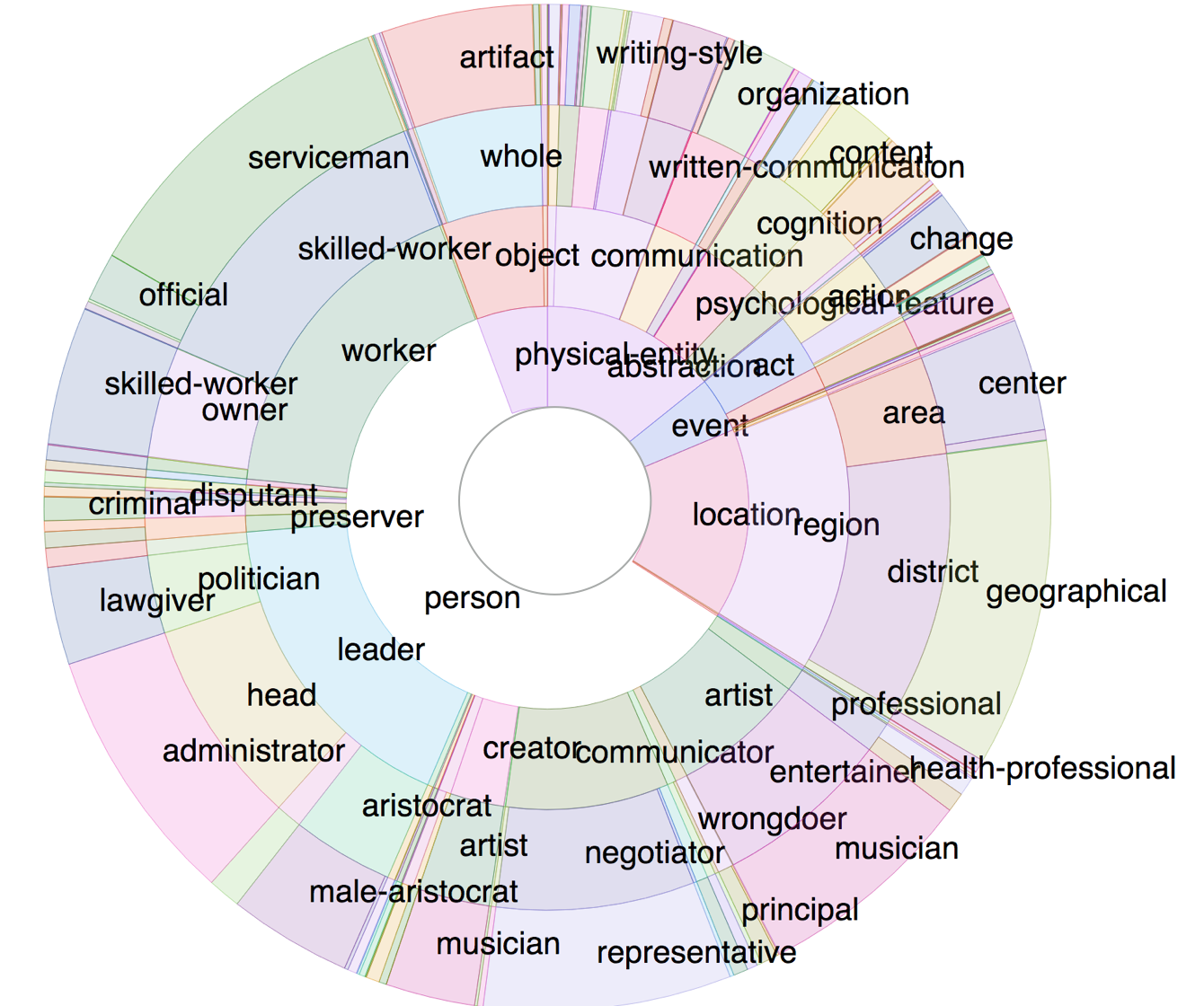}
\caption{Distribution of hierarchical WordNet synsets for entities appearing in the answer. The arc length is proportional to the number of questions containing that category.}
\label{fig:topic_dist}
\end{centering}
\end{figure}

\section{Dataset Analysis}
\label{sec:dataset_analysis}

A quantitative and qualitative analysis of TriviaQA shows it contains complex questions about a diverse set of entities, which are answerable using the evidence documents. 

\paragraph{Question and answer analysis}
TriviaQA questions, authored by trivia enthusiasts, cover various topics of people's interest. The average question length is 14 tokens indicating that many questions are highly compositional. For qualitative analysis, we sampled 200 question answer pairs and manually analysed their properties. About 73.5\% of these questions contain phrases that describe a fine grained category to which the answer belongs, while 15.5\% hint at a coarse grained category (one of \emph{person}, \emph{organization}, \emph{location}, and \emph{miscellaneous}). Questions often involve reasoning over time frames, as well as making comparisons. A summary of the analysis is presented in Table~\ref{tab:quest_analysis}. 



Answers in TriviaQA belong to a diverse set of types. 92.85\% of the answers are titles in Wikipedia,\footnote{This is a very large set since Wikipedia has more than 11 million titles.} 4.17\% are numerical expressions (e.g., 9 kilometres) while the rest are open ended noun and verb phrases. A coarse grained type analysis of answers that are Wikipedia entities presented in Table~\ref{tab:ans_analysis}. It should be noted that not all Wikipedia titles are named entities; many are common phrases such as \emph{barber}  or \emph{soup}. Figure~\ref{fig:topic_dist} shows diverse topics indicated by WordNet synsets of answer entities. 



\begin{table}[t]
\centering
\small
\begin{tabular}{l|r}\toprule
Type & Percentage \\\midrule
Numerical & 4.17\\ 
Free text & 2.98 \\
Wikipedia title & 92.85 \\ 
\hspace*{1em} Person & 32\\
\hspace*{1em} Location & 23\\
\hspace*{1em} Organization & 5   \\
\hspace*{1em} Misc. & 40\\ \toprule
\end{tabular}
\caption{Distribution of answer types on 200 annotated examples. }
\label{tab:ans_analysis}
\end{table}

\begin{table*}[t]
\small
\tabcolsep=0.11cm
\begin{center}
\begin{tabular}{l|cl}\toprule

Reasoning & \multicolumn{2}{l}{\textbf{Lexical variation (synonym)}}\\
 &\multicolumn{2}{l}{Major correspondences between the question and the answer sentence are synonyms.}\\
Frequency & \multicolumn{2}{l}{41\% in Wiki documents, 39\% in web documents.}\\
& Q& What is solid CO2 \underline{commonly called}? \\ 
\multirow{2}{*}{Examples} & S& The frozen solid form of CO2, \underline{known as} \textbf{dry ice} ... \\
  & Q& Who wrote the \underline{novel} The Eagle Has landed?\\
& S& The Eagle Has Landed is a \underline{book} by British writer \textbf{Jack Higgins}\\\midrule
{Reasoning}&\multicolumn{2}{l}{\textbf{Lexical variation and world knowledge}} \\
{} & \multicolumn{2}{l}{Major correspondences between the question and the document require common sense or external knowledge.}\\
Frequency &\multicolumn{2}{l}{17\% in Wiki documents, 17\% in web documents.}\\
&Q& What is the \underline{first name} of Madame Bovary in Flaubert's 1856 novel?\\
&  S&Madame Bovary (1856) is the French writer Gustave Flaubert's debut novel. The story focuses on a doctor's \\
Examples&& wife, \textbf{Emma} Bovary\\ 
&Q& Who was the \underline{female member} of the 1980's pop music duo, Eurythmics?\\ 
&S& Eurythmics were a British music duo consisting of members \textbf{Annie Lennox} and David A. Stewart. \\\midrule
{Reasoning}&\multicolumn{2}{l}{\textbf{Syntactic Variation}} \\
{} & \multicolumn{2}{l}{After the question is paraphrased into declarative form, its syntactic dependency structure does not match}\\
&\multicolumn{2}{l}{  that of the answer sentence}\\
Frequency &\multicolumn{2}{l}{69\% in Wiki documents, 65\% in web documents.}\\
&Q& In which country did the Battle of El Alamein take place?\\
\multirow{2}{*}{Examples}&S& The 1942 Battle of El Alamein in \textbf{Egypt} was actually two pivotal battles of World War II \\
&Q&Whom was Ronald Reagan referring to when he uttered the famous phrase evil empire in a 1983 speech?\\
&S&The phrase evil empire was first applied to the \textbf{Soviet Union} in 1983 by U.S. President Ronald Reagan.\\\midrule
{Reasoning}&\multicolumn{2}{l}{\textbf{Multiple sentences}} \\
&\multicolumn{2}{l}{Requires reasoning over multiple sentences.}\\
Frequency &\multicolumn{2}{l}{40\% in Wiki documents, 35\% in web documents.}\\
&Q&Name the Greek Mythological hero who killed the gorgon Medusa.\\
&S&\textbf{Perseus} asks god to aid him. So the goddess Athena and Hermes helps him out to kill Medusa.\\
Examples&Q&Who starred in and directed the 1993 film A Bronx Tale?\\
&S&\textbf{Robert De Niro} To Make His Broadway Directorial Debut With A Bronx Tale: The Musical. The actor\\
&& starred and directed the 1993 film.\\\midrule
Reasoning & \multicolumn{2}{l}{\textbf{Lists, Table}}\\
&\multicolumn{2}{l}{Answer found in tables or lists}\\
Frequency &\multicolumn{2}{l}{7\% in web documents.}\\
\multirow{2}{*}{Examples}&Q&In Moh's Scale of hardness, Talc is at number 1, but what is number 2?\\
&Q&What is the collective name for a group of hawks or falcons?\\ \bottomrule
\end{tabular}
\end{center}
\caption{Analysis of reasoning used to answer TriviaQA questions shows that a high proportion of evidence sentence(s) exhibit syntactic and lexical variation with respect to questions. Answers are indicated by boldfaced text. }
\label{tab:evidence_manual}
\end{table*}

\paragraph{Evidence analysis}
\label{sec:evidence_analysis}
A qualitative analysis of TriviaQA shows that the evidence contains answers for 79.7\%  and 75.4\% of questions from the Wikipedia and Web domains respectively. To analyse the quality of evidence and evaluate baselines, we asked a human annotator to answer 986 and 1345 (dev and test set) questions from the Wikipedia and Web domains respectively.  Trivia questions contain multiple clues about the answer(s) not all of which are referenced in the documents. The annotator was asked to answer a question if the minimal set of facts (ignoring temporal references like \emph{this year}) required to answer the question are present in the document, and abstain otherwise. For example, it is possible to answer the question, \emph{Who became president of the Mormons in 1844, organised settlement of the Mormons in Utah 1847 and founded Salt Lake City?} using only the fact that Salt Lake City was founded by Brigham Young. We found that the accuracy (evaluated using the original answers) for the Wikipedia and Web domains was 79.6 and 75.3 respectively. We use the correctly answered questions (and documents) as verified sets for evaluation (section \ref{experiments}).

\paragraph{Challenging problem}
A comparison of evidence with respect to the questions shows a high proportion of questions require reasoning over multiple sentences. To compare our dataset against previous datasets, we classified 100 question-evidence pairs each from Wikipedia and the Web according to the form of reasoning required to answer them. We focus the analysis on Wikipedia since the analysis on Web documents are similar. Categories are not mutually exclusive: single example can fall into multiple categories. A summary of the analysis is presented in Table~\ref{tab:evidence_manual}.

On comparing evidence sentences with their corresponding questions, we found that 69\% of the questions had a different syntactic structure while 41\% were lexically different. For 40\% of the questions, we found that the information required to answer them was scattered over multiple sentences. Compared to SQuAD, over three times as many questions in TriviaQA require reasoning over multiple sentences. Moreover, 17\% of the examples required some form of world knowledge. Question-evidence pairs in TriviaQA display more lexical and syntactic variance than SQuAD. This supports our earlier assertion that decoupling question generation from evidence collection results in a more challenging problem.

%% file: methods.tex
\section{Baseline methods}\label{sec:methods}
To quantify the difficulty level of the dataset for current methods, we present results on neural and other models. We used a random entity baseline and a simple classifier inspired from previous work \cite{WangMCT2015, chen2016thorough}, and compare these to BiDAF \cite{SeoKFH16},  one of the best performing models for the SQuAD dataset.

\subsection{Random entity baseline}
We developed the random entity baseline for the Wikipedia domain since the provided documents can be directly mapped to candidate answers. In this heuristic approach, we first construct a candidate answer set using the entities associated with the provided Wikipedia pages for a given question (on average 1.8 per question). We then randomly pick a candidate that does not occur in the question. If no such candidate exists, we pick any random candidate from the candidate set.

\subsection{Entity classifier}
We also frame the task as a ranking problem over candidate answers in the documents. More formally, given a question $q_i$, an answer $a_i^+$, and a evidence document $D_i$, we want to learn a scoring function $score$, such that
\begin{equation*}
score(a_i^+ | q_i, D_i) > score(a_i^- | q_i, D_i) 
\end{equation*}
\noindent where $a_i^-$ is any candidate other than the answer. The function $score$ is learnt using LambdaMART ~\cite{WuMart},\footnote{We use the RankLib implementation \url{https://sourceforge.net/p/lemur/wiki/RankLib/}} a boosted tree based ranking algorithm. 
 
This is similar to previous entity-centric classifiers for QA~\cite{chen2016thorough,WangMCT2015}, and uses context and Wikipedia catalog based features. To construct the candidate answer set, we consider sentences that contain at least one word in common with the question. We then add every n-gram ($n \in [1, 5]$) that occurs in these sentences and is a title of some Wikipedia article.\footnote{Using a named entity recognition system to generate candidate entities is not feasible as answers can be common nouns or phrases.} 

\subsection{Neural model} 
Recurrent neural network models (RNNs)~\cite{Hermann:15, chen2016thorough} have been very effective for reading comprehension. 
For our task, we modified the BiDAF model~\cite{SeoKFH16}, which takes a sequence of context words as input and outputs the start and end positions of the predicted answer in the context. The model utilizes an RNN at the character level, token level, and phrase level to encode context and question and uses attention mechanism between question and context.

Authored independently from the evidence document, TriviaQA does not contain the exact spans of the answers. We approximate the answer span by finding the first match of answer string in the evidence document. Developed for a dataset where the evidence document is a single paragraph (average 122 words), the BiDAF model does not scale to long documents. To overcome this, we truncate the evidence document to the first 800 words.\footnote{We found that splitting documents into smaller sub documents degrades performance since a majority of sub documents do not contain the answer.} 

When the data contains more than one evidence document, as in our Wikipedia domain, we predict for each document separately and aggregate the predictions by taking a sum of confidence scores. More specifically, when the model outputs a candidate answer $A_i$ from $n$ documents $D_{i, 1}, ... D_{i, n}$ with confidences ${c_{i, 1}, ... c_{i, n} }$, the score of $A_i$ is given by
    \begin{equation*}
    score(A_i) = \sum_k c_{i,k}   
    \end{equation*}
We select candidate answer with the highest score.  

    

%% file: experiments.tex
\section{Experiments}
\label{experiments}
\begin{table}

\begin{center}
\small
\begin{tabular}{l|l|r|r|r}\toprule
\multicolumn{2}{c|}{}& Train & Dev & Test \\ \hline
\multirow{2}{*}{Wikipedia}&Questions & 61,888 & 7,993 & 7,701\\ 
&Documents & 110,648 & 14,229 & 13,661\\ \hline
\multirow{2}{*}{Web}&Questions & 76,496 & 9,951 & 9,509\\ 
&Documents & 528,979 & 68,621 & 65,059\\ \hline
\multirow{2}{*}{\parbox{1.5cm}{Wikipedia verified}}&Questions & - & 297 & 584\\ 
&Documents & - & 305 & 592\\ \hline
{Web }&Questions & - & 322 & 733\\ 
verified&Documents & - & 325 & 769\\ 

\bottomrule
\end{tabular}
\end{center} \vspace{-5pt}
\caption{Data statistics for each task setup. The Wikipedia domain is evaluated over questions while the web domain is evaluated over documents.}
\label{tab:task_setup}
\end{table}

\begin{table*}
\centering
\footnotesize
\tabcolsep=0.11cm
\begin{tabular}{c|c|c|c|c|c|c|c|c|c|c|c|c|c}\toprule
& & \multicolumn{6}{c|}{Distant Supervision}& \multicolumn{6}{c}{Verified} \\ 
{Method} & {Domain} &\multicolumn{3}{c|}{Dev} & \multicolumn{3}{c|}{Test}&\multicolumn{3}{c|}{Dev} & \multicolumn{3}{c}{Test} \\
& & EM & F1 & Oracle & EM & F1 & Oracle & EM & F1 & Oracle & EM & F1 & Oracle\\ \toprule 
Random &  & 12.72 & 22.91 & 16.30 & 12.74 & 22.35 & 16.28 & 14.81 & 23.31 & 19.53 & 15.41 & 25.44 & 19.19\\
Classifier & Wiki &  23.42 & 27.68 & 71.41 & 22.45  & 26.52 & 71.67 & 24.91 & 29.43 & 80.13 & 27.23 & 31.37 & 77.74\\
\textbf{BiDAF} &  & \textbf{40.26} & \textbf{45.74} & \textbf{82.55} & \textbf{40.32} & \textbf{45.91} & \textbf{82.82} & \textbf{47.47} & \textbf{53.70} & \textbf{90.23} & \textbf{44.86} & \textbf{50.71} & \textbf{86.81}\\
\toprule
Classifier& \multirow{2}{*}{web} & 24.64  & 29.08 & 66.78 &  24.00 & 28.38 & 66.35 & 27.38 & 31.91 & 77.23 & 30.17 & 34.67 & 76.72\\
\textbf{BiDAF} &  & \textbf{41.08} & \textbf{47.40} & \textbf{82.93}  & \textbf{40.74} & \textbf{47.05} & \textbf{82.95} & \textbf{51.38} & \textbf{55.47} & \textbf{90.46} & \textbf{49.54} & \textbf{55.80} & \textbf{89.99}\\ \bottomrule
\end{tabular}

\caption{Performance of all systems on TriviaQA using distantly supervised evaluation. The best performing system is indicated in bold.}
\label{tab:main_results}
\end{table*}

An evaluation of our baselines shows that both of our tasks are challenging, and that the TriviaQA dataset supports significant future work.

\label{sec:eval_metrics}
\subsection{Evaluation Metrics}
We use the same evaluation metrics as SQuAD -- exact match (EM) and F1 over words in the answer(s). For questions that have \emph{Numerical} and \emph{FreeForm} answers, we use a single given answer as ground truth. For questions that have Wikipedia entities as answers, we use Wikipedia aliases as valid answer along with the given answer.

Since Wikipedia and the web are vastly different in terms of style and content, we report performance on each source separately. While using Wikipedia, we evaluate at the question level since facts needed to answer a question are generally stated only once. On the other hand, due to high information redundancy in web documents (around 6 documents per question), we report document level accuracy and F1 when evaluating on web documents. Lastly, in addition to distant supervision, we also report evaluation on the clean dev and test questions collection using a human annotator (section \ref{sec:evidence_analysis})

\label{sec:expt_setup}
\subsection{Experimental Setup}
We randomly partition QA pairs in the dataset into train (80\%), development (10\%), and test set (10\%). In addition to distant supervision evaluation, we also evaluate baselines on verified subsets (see section \ref{sec:evidence_analysis}) of the dev and test partitions. Table \ref{tab:task_setup} contains the number of questions and documents for each task. We trained the entity classifier on a random sample of 50,000 questions from the training set. For training BiDAF on the web domain, we first randomly sampled 80,000 documents. For both domains, we used only those (training) documents where the answer appears in the first 400 tokens to keep training time manageable. Designing scalable techniques that can use the entirety of the data is an interesting direction for future work.






\subsection{Results}
The performance of the proposed models is summarized in Table~\ref{tab:main_results}. The poor performance of the random entity baseline shows that the task is not already solved by information retrieval. For both Wikipedia and web documents, BiDAF (40\%) outperforms the classifier (23\%). The oracle score is the upper bound on the exact match accuracy.\footnote{A question $q$ is considered answerable for the oracle score if the correct answer is found in the evidence $D$ or, in case of the classifier, is a part of the candidate set. Since we truncate documents, the upper bound is not 100\%.} All models lag significantly behind the human baseline of 79.7\% on the Wikipedia domain, and 75.4\% on the web domain.

We analyse the performance of BiDAF on the development set using Wikipedia as the evidence source by question length and answer type. The accuracy of the system steadily decreased as the length of the questions increased -- with 50\% for questions with 5 or fewer words to 32\% for 20 or more words. This suggests that longer compositional questions are harder for current methods. 



\subsection{Error analysis}
Our qualitative error analysis reveals that compositionality in questions and lexical variation and low signal-to-noise ratio in (full) documents is still a challenge for current methods. We randomly sampled 100 incorrect BiDAF predictions from the development set and used Wikipedia evidence documents for manual analysis. We found that 19 examples lacked evidence in any of the provided documents, 3 had incorrect ground truth, and 3 were valid answers that were not included in the answer key. Furthermore, 12 predictions were partially correct (\emph{Napoleonic} vs \emph{Napoleonic Wars}). This seems to be consistent with human performance of 79.7\%.

For the rest, we classified each example into one or more categories listed in Table \ref{tab:qual_error_analysis}. Distractor entities refers to the presence of entities similar to ground truth. E.g., for the question, \emph{Rebecca Front plays Detective Chief Superintendent Innocent in which TV series?}, the evidence describes all roles played by Rebecca Front. 

The first two rows suggest that long and noisy documents make the question answering task more difficult, as compared for example to the short passages in SQuAD. Furthermore, a high proportion of errors are caused by paraphrasing, and the answer is sometimes stated indirectly. For example, the evidence for the question \emph{What was Truman Capote's last name before he was adopted by his stepfather?} consists of the following text \emph{Truman Garcia Capote born Truman Streckfus Persons, was an American ... In 1933, he moved to New York City to live with his mother and her second husband, Joseph Capote, who adopted him as his stepson and renamed him Truman García Capote.}

\begin{table}[t]
\small
    \centering
    \begin{tabular}{c|c}
    \toprule
    Category & Proportion \\ \midrule
    Insufficient evidence & 19 \\
     Prediction from incorrect document(s) & 7  \\
     Answer not in clipped document & 15\\
     Paraphrasing & 29\\
     Distractor entities  & 11\\
     Reasoning over multiple sentences & 18\\
         \toprule
    \end{tabular}
    \caption{Qualitative error analysis of BiDAF on Wikipedia evidence documents.}
    \label{tab:qual_error_analysis}
\end{table}



%% file: existingDatasets.tex
\section{Related work}
Recent interest in question answering has resulted in the creation of several datasets. However, they are either limited in scale or suffer from biases stemming from their construction process. We group existing datasets according to their associated tasks, and compare them against TriviaQA. The analysis is summarized in Table~\ref{tab:dataset_comparison}.

\subsection{Reading comprehension}
Reading comprehension tasks aims to test the ability of a system to understand a document using questions based upon its contents. Researchers have constructed cloze-style datasets ~\cite{Hill:15,Hermann:15,Paperno:16,Onishi:16}, where the task is to predict missing words, often entities, in a document. Cloze-style datasets, while easier to construct large-scale  automatically
, do not contain natural language questions. 

Datasets with natural language questions include MCTest \cite{Richardson:13}, SQuAD ~\cite{Rajpurkar:16}, and NewsQA \cite{Trischler:16}. MCTest is limited in scale with only 2640 multiple choice questions. SQuAD contains 100K crowdsourced questions and answers paired with short Wikipedia passages. NewsQA uses crowdsourcing to create questions solely from news article summaries in order to control potential bias. The crucial difference between SQuAD/NewsQA and TriviaQA is that TriviaQA questions have not been crowdsourced from pre-selected passages. Additionally, our evidence set consists of web documents, while SQuAD and NewsQA are limited to Wikipedia  and news articles respectively. Other recently released datasets include \cite{lai2017large}.

\subsection{Open domain question answering}

The recently released MS Marco dataset~\cite{Nguyen:16} also contains independently authored questions and documents drawn from the search results. However, the questions in the dataset are derived from search logs and the answers are crowdsourced. On the other hand, trivia enthusiasts provided both questions and answers for our dataset.

Knowledge base question answering involves converting natural language questions to logical forms that can be executed over a KB. Proposed datasets~\cite{CaiY13,BerantWebQ2013,BordesSimpleQ2015} are either limited in scale or in the complexity of questions, and can only retrieve facts covered by the KB. 



A standard task for open domain IR-style QA is the annual TREC competitions~\cite{Voorhees:00}, which contains questions from various domains but is limited in size. 
Many advances from the TREC competitions were used in the IBM Watson system for \emph{Jeopardy!} ~\cite{Ferrucci:10}. Other datasets includes SearchQA \cite{searchqa} where \emph{Jeopardy!} questions are paired with search engine snippets,  
the WikiQA dataset \cite{WikiQA:16} 
 for answer sentence selection, and the Chinese language WebQA \cite{LiLHWCZXWebQA16} dataset, 
which focuses on the task of answer phrase extraction. TriviaQA contains examples that could be used for both stages of the pipeline, although our focus on this paper is instead on using the data for reading comprehension where the answer is always present.

Other recent approaches attempt to combine structured high precision KBs with semi-structured information sources like OpenIE triples \cite{Fader14}, HTML tables \hbox{\cite{Pasupat2015CompositionalSP}}, and large (and noisy) corpora ~\cite{SawantC13,JoshiEMNLP2014,Xu:15}. TriviaQA, which has Wikipedia entities as answers, makes it possible to leverage structured KBs like Freebase, which we leave to future work. Furthermore, about 7\% of the TriviaQA questions have answers in HTML tables and lists, which could be used to augment these existing resources. 

Trivia questions from quiz bowl have been previously used in other question answering tasks \cite{boydgraber2012}.  Quiz bowl questions are paragraph length and pyramidal.\footnote{Pyramidal questions consist of a series of clues about the answer arranged in order from most to least difficult. } A number of different aspects of this problem have been carefully studied, typically using classifiers over a pre-defined set of answers~\cite{Iyyer:14} and studying incremental answering to answer as quickly as possible~\cite{boydgraber2012} or using reinforcement learning to model opponent behavior~\cite{he2016opponent}. These competitive challenges are not present in our single-sentence question setting. Developing joint models for multi-sentence reasoning for questions and answer documents is an important area for future work.

%% file: conclusion.tex
\section{Conclusion and Future Work}
We present TriviaQA, a new dataset of 650K question-document-evidence triples. To our knowledge, TriviaQA is the first dataset where questions are authored by trivia enthusiasts, independently of the evidence documents. The evidence documents come from two domains -- Web search results and Wikipedia pages -- with highly  differing  levels  of  information  redundancy. Results from current state-of-the-art baselines indicate that TriviaQA is a challenging testbed that deserves significant future study.

While not the focus of this paper, TriviaQA also provides a provides a benchmark for a variety of other tasks such as IR-style question answering, QA over structured KBs and joint modeling of KBs and text, with much more data than previously available. 

%% file: ack.tex
\section*{Acknowledgments}
This work was supported by DARPA contract FA8750-13-2-0019, the WRF/Cable Professorship, gifts from Google and Tencent, and an Allen Distinguished Investigator Award.
The authors would like to thank Minjoon Seo for the BiDAF code, and Noah Smith, Srinivasan Iyer, Mark Yatskar,  Nicholas FitzGerald, Antoine Bosselut, Dallas Card, and anonymous reviewers for helpful comments.